\documentclass[10pt,twocolumn,letterpaper]{article}

\usepackage[pagenumbers]{cvpr}

\definecolor{cvprblue}{rgb}{0.21,0.49,0.74}
\usepackage[pagebackref,breaklinks,colorlinks,allcolors=cvprblue]{hyperref}

\def\confName{CVPR}
\def\confYear{2027}

\title{MIRA: Multi-dimensional Reward Alignment for Hand-Object Interaction Video Generation}

\author{Zhenhao Zhang$^1$
\quad  Dengpeng Wu$^1$
\quad Lingwei Dang$^{1,2}$
\quad Jiajun Zhang$^1$
\quad Wei MIN$^3$
\quad Yebin Liu$^{1,}$
\thanks{Corresponding author.} \\
$^1$Tsinghua University
$^2$South China University of Technology
$^3$Shadow AI
\\
}

\makeatletter
\g@addto@macro\@maketitle{%
  \noindent
  \begin{minipage}{\textwidth}
    \centering
    \captionsetup{type=figure}

    \includegraphics[width=\linewidth]{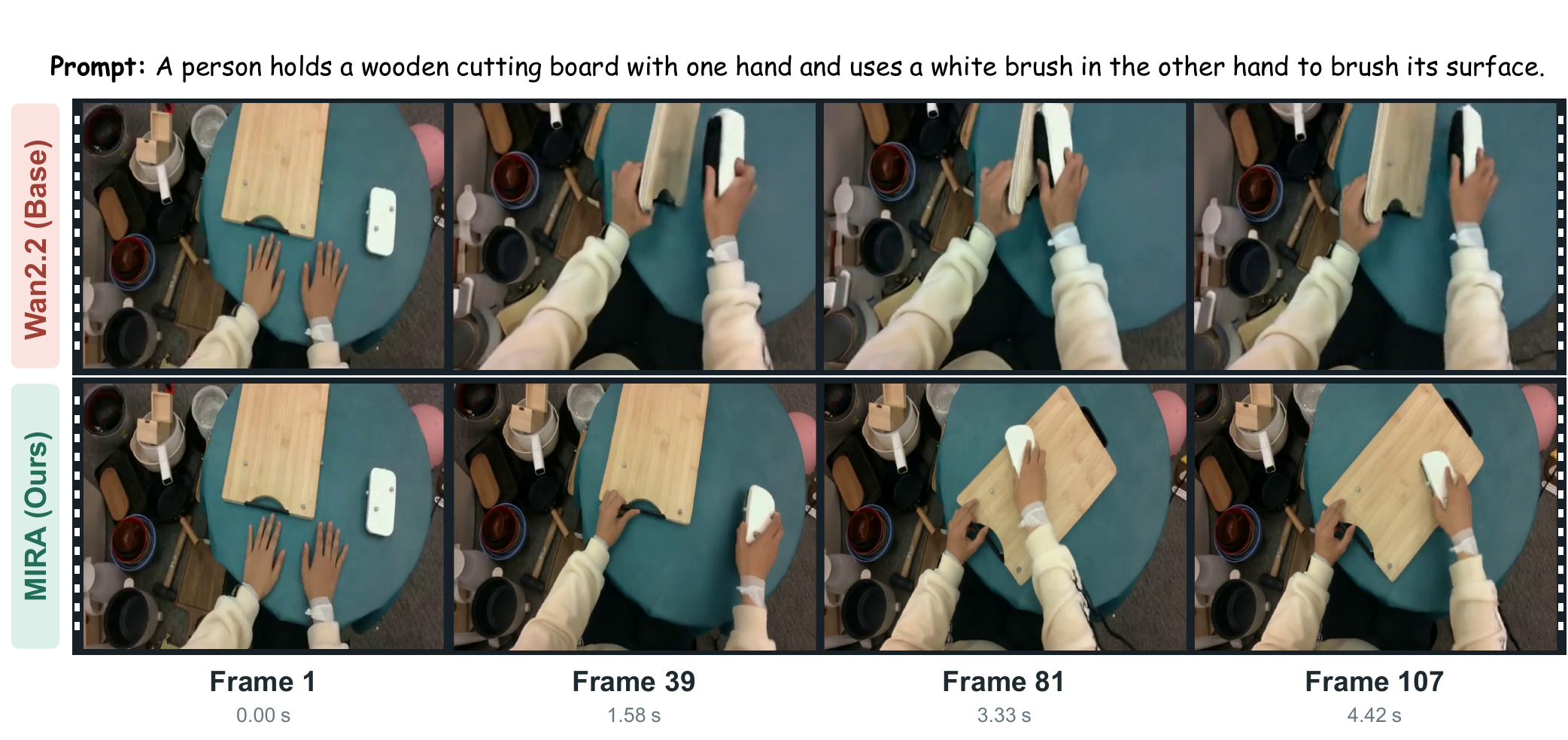}

    \caption{%
      \textbf{Comparison with the base model.}
      Given the same initial image, Wan2.2 (top) and MIRA (bottom) generate videos of brushing a wooden cutting board. Wan2.2 produces unnatural bending and distortions in the outline of the board as it is lifted and manipulated. MIRA better preserves the rigid shape of the board throughout the interaction.
    }
    \label{fig:teaser}
  \end{minipage}\par
  \vspace{1em}
}
\makeatother
\begin{document}
\maketitle
\begin{abstract}
Hand-object interaction (HOI) video generation requires visual quality, 3D spatial consistency, and physically plausible motion. Existing methods often rely on additional interaction conditions that are difficult to obtain at inference time, while RL-based approaches lack effective constraints on 3D interactions and dynamic scenes. To address these limitations, we propose \textbf{MIRA}, to the best of our knowledge, the first RL post-training framework for HOI video generation that jointly leverages 2D, 3D, and 4D rewards. Specifically, the 2D reward evaluates visual quality and semantic consistency, the 3D reward measures cross-view geometric consistency, and the 4D reward captures hand motion consistency and dynamic scene-structure consistency over time. Furthermore, we introduce rank-based preference assignment and multi-dimensional consensus filtering to construct high-confidence preference pairs for RL post-training. Experiments across multiple datasets demonstrate that MIRA achieves state-of-the-art performance on test splits without additional inference-time overhead.
\end{abstract}    
\section{Introduction}
Hand-object interaction (HOI) video generation is an important research topic in computer vision~\cite{svimo,chen2026hvg,dang2026harmohoi} and digital human synthesis~\cite{openhoi,bimart,diffh2o}, with broad applications in AR/VR~\cite{qian2022arnnotate,chen2026ubi,cho2026unified} and embodied AI~\cite{zhang2026unihm,li2026scalable,luo2025being}. However, generating HOI videos with smooth motion, visual realism, and physical plausibility remains challenging.

Existing HOI video generation methods can be broadly categorized into conditioning-based and joint-generation approaches. Conditioning-based methods~\cite{chen2026mimic,chen2026hvg,chen2026handsonworld} use additional hand motion sequences or interaction representations to guide video synthesis, while joint-generation methods~\cite{svimo,dang2026harmohoi,scar} employ multimodal architectures to jointly generate videos and HOI representations. Despite their progress, conditioning-based methods rely on additional interaction inputs that are difficult to obtain at inference time, whereas joint-generation methods increase architectural and training complexity and may interfere with the generative priors of pretrained video models.

Recent RL-based post-training methods provide an effective way to improve video generation quality without requiring additional interaction inputs or inference-time computational overhead, as reward models are only used during training. Existing approaches mainly follow two directions. One line of work~\cite{videodpo,cai2026phygdpo} employs pretrained vision models to evaluate generated videos, providing feedback on visual quality and physical properties, but lacks explicit constraints on 3D spatial relationships and cross-view geometric consistency. Another line~\cite{du2026videogpa,worldr1} leverages 3D reconstruction to provide geometric feedback, but primarily focuses on static scenes, limiting its ability to model dynamic spatial relationships and motion in hand-object interactions. Therefore, jointly evaluating visual quality, cross-view geometric consistency, and dynamic HOI motion without additional inference-time overhead remains challenging.

To address these limitations, we propose \textbf{MIRA}, to the best of our knowledge, the first RL post-training framework specifically designed for HOI video generation, which jointly leverages 2D, 3D, and 4D rewards to evaluate generated videos. Specifically, the 2D reward evaluates visual quality and semantic consistency, the 3D reward assesses 3D geometry and cross-view consistency, and the 4D reward captures hand motion consistency and temporal consistency of dynamic scene structures.
To accommodate the different numerical scales of these rewards, we therefore introduce rank-based preference assignment, which independently ranks candidate videos within each reward dimension and converts their relative rankings into preference labels. Multi-dimensional consensus filtering then selects preferred and dispreferred samples supported by multiple dimensions without conflicting votes. This procedure integrates complementary reward judgments into reliable preference pairs for RL post-training.

We conduct extensive experiments on multiple datasets. Results demonstrate that MIRA achieves state-of-the-art performance on test splits, outperforming existing methods in video quality and hand pose accuracy. Furthermore, under the same training budget, consensus-filtered preference data yields better generation performance than the unfiltered data, validating the effectiveness of our data filtering strategy. Notably, these improvements are achieved without additional inference-time computational overhead.

Our main contributions are summarized as follows:

\begin{itemize}
    \item We propose \textbf{MIRA}, the first RL post-training framework for HOI video generation, which improves generation quality without requiring additional interaction inputs or computational overhead at inference time.

    \item We develop a unified 2D, 3D, and 4D reward modeling framework, together with rank-based preference assignment and multi-dimensional consensus filtering, to construct high-confidence preference data for RL post-training.

    \item Extensive experiments across multiple datasets demonstrate that MIRA achieves state-of-the-art performance on test splits, generating HOI videos with smoother motion, stronger spatial consistency, and greater physical plausibility.
\end{itemize}

\begin{figure*}[t]
    \centering
    \includegraphics[width=\textwidth]{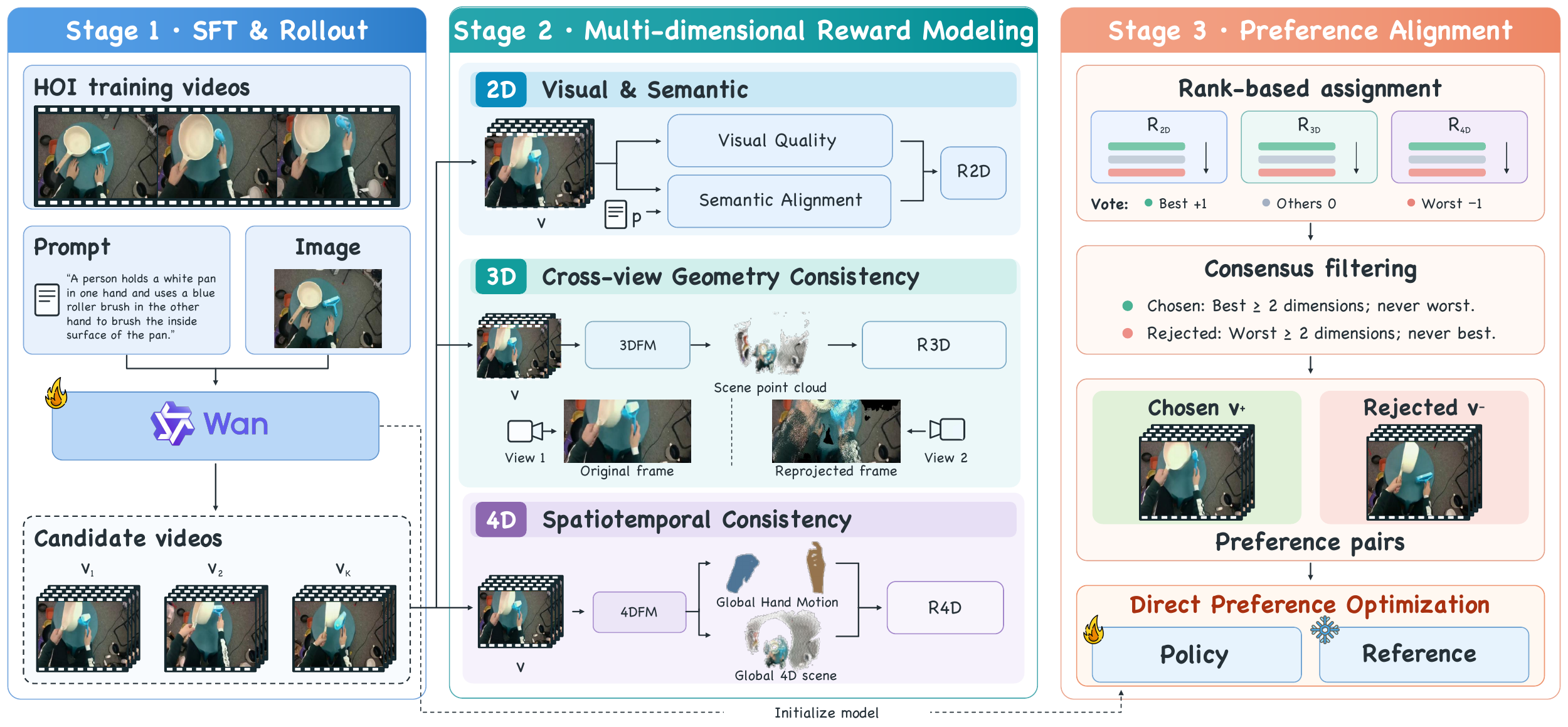}

    \caption{\textbf{Overview of MIRA.}
    We first perform supervised fine-tuning on HOI videos and
    use the SFT model to generate candidate videos conditioned on initial images and text prompts. These candidates are evaluated with 2D, 3D, and 4D rewards.
    Rank-based preference assignment and multi-dimensional
    consensus filtering construct reliable preference pairs,
    which are then used for DPO post-training of the video
    generation model.}

    \label{fig:method}
\end{figure*}
\section{Related Works}

\paragraph{Hand-object Interaction Motion Generation.}
For 3D motion generation, Text2HOI~\cite{text2hoi} and DiffH2O~\cite{diffh2o} synthesize hand-object motions conditioned on textual descriptions and object geometry. ManiDext~\cite{manidext} and BimArt~\cite{bimart} further incorporate object trajectories and contact priors to model dexterous manipulation, while OpenHOI~\cite{openhoi} and HOIGPT~\cite{hoigpt} explore large language model-based approaches for 3D HOI generation. Although these methods explicitly model hand-object motions and interaction relationships, they primarily focus on 3D motion synthesis rather than generating visually rich videos.

\paragraph{Hand-object Interaction Video Generation.}
For HOI video generation, ManiVideo~\cite{manivideo} and HVG-3D~\cite{chen2026hvg} use external 3D interaction sequences as motion conditions, while MIMIC~\cite{chen2026mimic} employs interaction masks to guide video synthesis. Alternatively, SViMo~\cite{svimo}, HarmoHOI~\cite{dang2026harmohoi}, and SCAR~\cite{scar} model videos and interaction representations via joint generation or collaborative denoising. Despite their progress, conditioning-based methods rely on additional interaction inputs at inference time, whereas joint-generation methods introduce auxiliary representations and multimodal modeling, increasing architectural complexity and inference-time computational overhead.

\paragraph{Video Generation with Reinforcement Learning.}
RL-based post-training provides an effective way to improve video generation quality. VideoDPO~\cite{videodpo} and PhyGDPO~\cite{cai2026phygdpo} leverage visual and physics rewards while DanceGRPO~\cite{xue2025dancegrpo} applies group relative
policy optimization using visual and motion
quality feedback. However, these approaches lack explicit constraints on cross-view geometric consistency and HOI-specific motion dynamics. VideoGPA~\cite{du2026videogpa} and World-R1~\cite{worldr1} incorporate 3D reconstruction rewards but primarily rely on static-scene assumptions, limiting its applicability to dynamic HOI scenes. %In concurrent work, 
In contrast, MIRA jointly leverages 2D, 3D, and 4D rewards, together with rank-based preference assignment and consensus-based filtering, to construct high-confidence preference data for RL post-training.

\section{Method}

We propose \textbf{MIRA}, an RL post-training framework for hand-object interaction video generation. 
Given a text prompt $p$ and an initial image
$I_{\mathrm{init}}$, we denote the video generation condition
as $c=(p,I_{\mathrm{init}})$.
We evaluate generated videos using 2D, 3D, and 4D rewards, convert the scores into relative preferences, and select reliable preference pairs through multi-dimensional consensus. The resulting pairs are used to post-train the video generation model with Direct Preference Optimization. More details in the supplementary material.

\subsection{Multi-Dimensional Reward Modeling}

\subsubsection{2D Reward Model}

The 2D reward assesses visual quality\cite{li2023amt,teed2020raft,caron2021dino,ke2021musiq} and semantic alignment\cite{wang2024internvid}. Given a generated video $v$ and its text prompt $p$, we compute a visual quality score $Q(v)$ and a semantic alignment score $S(v,p)$. Details are provided in the supplementary material.

\paragraph{Visual Quality.}
We evaluate intra-frame quality using imaging and aesthetic scores, and inter-frame quality using dynamic degree, motion smoothness, temporal flickering, and subject consistency. The latter are measured using optical flow, frame interpolation, changes in static regions, and cross-frame feature similarity, respectively. After normalizing the six scores to a common higher-is-better direction, we compute:

\begin{equation}
    Q(v)=\frac{1}{|\mathcal{K}_Q|}
    \sum_{k\in\mathcal{K}_Q}\widetilde{s}_k(v),
    \qquad |\mathcal{K}_Q|=6,
    \label{eq:visual_quality}
\end{equation}

where $\mathcal{K}_Q$ is the set of visual quality metrics and $\widetilde{s}_k(v)$ is the normalized score of metric $k$.

\paragraph{Semantic Alignment.}
We measure text-video alignment using the similarity between video and text features in a shared embedding space. The normalized semantic score is:

\begin{equation}
    S(v,p)=\mathcal{N}_S\!\left(
    \operatorname{sim}\big(f_v(v),f_p(p)\big)\right),
    \label{eq:semantic_alignment}
\end{equation}

where $f_v$ and $f_p$ denote the video and text encoders, and $\mathcal{N}_S$ denotes score normalization.

\paragraph{Reward Aggregation.}
We combine the two components with a 4:1 weighting ratio:

\begin{equation}
    R_{\mathrm{2D}}(v,p)=\frac{4Q(v)+S(v,p)}{5}.
    \label{eq:2d_reward}
\end{equation}

A higher $R_{\mathrm{2D}}$ indicates better overall visual quality and semantic alignment.

\subsubsection{3D Reward Model}

The 3D reward evaluates cross-view geometry consistency across sampled frames through scene reconstruction and reprojection\cite{wang2025vggt}. For sampled frames of $v=\{I_t\}_{t=1}^{T}$, we estimate depth $D_t$, camera intrinsics $\mathbf K_t$, and camera-to-world pose $(\mathbf R_t,\mathbf t_t)$.

\paragraph{Geometric Reconstruction.}
We unproject pixels into 3D and transform them into world coordinates, then aggregate points and colors across frames into a shared scene representation $\mathcal P$:

\begin{equation}
\begin{aligned}
    \mathbf{x}_t(\mathbf q)
    &=D_t(\mathbf q)\mathbf K_t^{-1}\widetilde{\mathbf q},\\
    \mathbf X_t(\mathbf q)
    &=\mathbf R_t\mathbf{x}_t(\mathbf q)+\mathbf t_t,
\end{aligned}
\label{eq:3d_reconstruction}
\end{equation}

where $\mathbf q$ denotes the 2D pixel coordinate, $\widetilde{\mathbf q}$ is its homogeneous representation, and $\mathbf{x}_t$ and $\mathbf X_t$ are the camera- and world-space points, respectively.

\paragraph{Cross-view Consistency.}
We reproject $\mathcal P$ into each sampled view to obtain $\hat I_t$, using the projection $\pi\big(\mathbf K_t\mathbf R_t^\top(\mathbf X-\mathbf t_t)\big)$. Pixel and perceptual discrepancies are combined over the sampled frame set $\mathcal T_{\mathrm{3D}}$:

\begin{equation}
\begin{aligned}
 E_{\mathrm{3D}}(v)
 &=\frac{1}{T_{\mathrm{3D}}}
 \sum_{t\in\mathcal T_{\mathrm{3D}}}
 \big[\operatorname{MSE}(I_t,\hat I_t)\\
 &\hspace{4.0em}
 +\operatorname{LPIPS}(I_t,\hat I_t)\big],\\
 R_{\mathrm{3D}}(v)&=-E_{\mathrm{3D}}(v),
\end{aligned}
\label{eq:3d_reward}
\end{equation}

where $T_{\mathrm{3D}}=|\mathcal T_{\mathrm{3D}}|$. Lower reprojection error yields a higher 3D reward.

\subsubsection{4D Reward Model}

The 4D reward primarily evaluates spatiotemporal consistency of generated video. It evaluates HOI motion consistency by comparing the hand motion reconstructed\cite{zhu2026mint} from the generated video $v$ and reference videos $v^{\mathrm{gt}}$. It further incorporates dynamic scene-structure\cite{zhuo2026streaming} consistency to capture the temporal evolution of 4D environments during interactions.

\paragraph{Hand Motion Consistency.}
We reconstruct MANO\cite{romero2017embodied} sequences $\mathbf h_t$ from each frame of both videos, comprising the translation, orientation, articulation, and shape parameters of the two hands. The accumulated hand motion error is:

\begin{equation}
 E_{\mathrm H}(v,v^{\mathrm{gt}})
 =\sum_{t=1}^{T}
 \left\|
 \mathbf h_t(v)-\mathbf h_t(v^{\mathrm{gt}})
 \right\|_2.
 \label{eq:4d_hand_error}
\end{equation}

\paragraph{Dynamic Scene-Structure Consistency.}
We reconstruct dense 4D point maps $\mathbf P_t^v$ and $\mathbf P_t^{\mathrm{gt}}$ from the generated and reference videos, respectively, using streaming geometry estimation. Since the two videos are reconstructed independently, their 4D point maps may lie in different coordinate systems. We therefore estimate a single video-level similarity transformation to align the generated reconstruction with the reference:

\begin{equation}
    \mathcal A_v(\mathbf X)
    =
    s_v\mathbf R_v^a\mathbf X
    +
    \mathbf t_v^a,
    \label{eq:video_alignment}
\end{equation}

where $s_v$, $\mathbf R_v^a$, and $\mathbf t_v^a$ denote the scale factor, rotation matrix, and translation vector, respectively. The same transformation is applied to all frames to preserve the relative spatial relationships and motion variations throughout the video.

Based on the aligned reconstructions, we calculate the frame-wise 3D discrepancies at corresponding pixel locations and aggregate them across the entire video:

\begin{equation}
\begin{aligned}
 E_{\mathrm G}(v,v^{\mathrm{gt}})
 &=\sum_{t=1}^{T}\frac{1}{|\Omega_t|}
 \sum_{\mathbf q\in\Omega_t}
 \Big\|
 \mathcal A_v\big(\mathbf P_t^v(\mathbf q)\big)\\
 &\qquad\qquad
 -\mathbf P_t^{\mathrm{gt}}(\mathbf q)
 \Big\|_2,
\end{aligned}
\label{eq:4d_geometry_error}
\end{equation}

where $\Omega_t$ denotes the set of valid pixel correspondences in the $t$-th frame. This error measures dynamic scene-structure consistency by comparing frame-wise 3D point positions in a shared coordinate system. A lower error indicates greater consistency between the generated and reference videos.

\paragraph{4D Reward Aggregation.}
For candidates $\mathcal V_c$ , we rank $E_{\mathrm H}$ and $E_{\mathrm G}$ in ascending order, where lower errors receive better ranks. Their ranks $r_{\mathrm H}$ and $r_{\mathrm G}$ are combined as:

\begin{equation}
\begin{aligned}
 \rho_{\mathrm{4D}}(v)
 &=r_{\mathrm H}(v)+r_{\mathrm G}(v),\\
 R_{\mathrm{4D}}(v\mid\mathcal V_c,v^{\mathrm{gt}})
 &=-\rho_{\mathrm{4D}}(v).
\end{aligned}
\label{eq:4d_reward}
\end{equation}

For candidates with identical rank sums, we use the normalized joint error as a secondary criterion:

\begin{equation}
\begin{aligned}
 J_{\mathrm{4D}}(v)
 &=\frac{E_{\mathrm H}(v,v^{\mathrm{gt}})-E_{\mathrm H}^{\min}}
 {\Delta_{\mathrm H}}\\
 &\quad+
 \frac{E_{\mathrm G}(v,v^{\mathrm{gt}})-E_{\mathrm G}^{\min}}
 {\Delta_{\mathrm G}},
\end{aligned}
\label{eq:4d_tie_breaking}
\end{equation}

where $E_{\mathrm H}^{\min}$ and $E_{\mathrm G}^{\min}$ are the minimum errors within the current candidate group, and $\Delta_{\mathrm H}$ and $\Delta_{\mathrm G}$ denote the median within-group error ranges across candidate groups. A lower $J_{\mathrm{4D}}$ indicates a better candidate.

The resulting 4D reward is subsequently used for rank-based preference assignment.

\subsection{Preference Pair Construction}

We convert reward scores into within-group preference labels and filter the resulting pairs using agreement across the three reward dimensions.

\subsubsection{Rank-based Preference Assignment}

For each text prompt $p$ and first frame condition $I_{\mathrm{init}}$, we generate $K$ videos $\mathcal V_c=\{v_1,\ldots,v_K\}$ using different random seeds. For each candidate $v_i$, we compute the 2D reward $R_{\mathrm{2D}}(v_i,p)$, the 3D reward $R_{\mathrm{3D}}(v_i)$, and the 4D reward $R_{\mathrm{4D}}(v_i\mid\mathcal V_c,v^{\mathrm{gt}})$. All three rewards are defined such that higher values indicate better evaluation results.

For each reward dimension $d\in\{\mathrm{2D},\mathrm{3D},\mathrm{4D}\}$, we rank the candidates in descending order of their corresponding reward scores. The highest- and lowest-ranked candidates are selected as the preferred and dispreferred videos, respectively:

\begin{equation}
\begin{aligned}
 v_d^+
 &=\underset{v_i\in\mathcal V_c}{\arg\max}\;R_d(v_i),\\
 v_d^-
 &=\underset{v_i\in\mathcal V_c}{\arg\min}\;R_d(v_i),\\
 \ell_d(v_i)&=
 \begin{cases}
 +1,&v_i=v_d^+,\\
 -1,&v_i=v_d^-,\\
 0,&\text{otherwise}.
 \end{cases}
\end{aligned}
\label{eq:rank_preference}
\end{equation}

Here $R_d(v_i)$ denotes the reward of dimension $d$, with its conditioning variables omitted for clarity, corresponding to $R_{\mathrm{2D}}(v_i,p)$, $R_{\mathrm{3D}}(v_i)$, or $R_{\mathrm{4D}}(v_i\mid\mathcal V_c,v^{\mathrm{gt}})$. The preference label $\ell_d(v_i)\in\{-1,0,+1\}$ represents rejected, unselected, and chosen, respectively. Ranking avoids directly combining rewards with different numerical scales.

\subsubsection{Consensus-based Data Filtering}

The reward dimensions may disagree on a candidate. We therefore count its chosen and rejected votes and retain only candidates supported by a majority without conflicting votes:

\begin{equation}
\begin{aligned}
 N_+(v_i)&=\sum_{d\in\{\mathrm{2D},\mathrm{3D},\mathrm{4D}\}}
 \mathbb I[\ell_d(v_i)=+1],\\
 N_-(v_i)&=\sum_{d\in\{\mathrm{2D},\mathrm{3D},\mathrm{4D}\}}
 \mathbb I[\ell_d(v_i)=-1].
\end{aligned}
\label{eq:consensus_votes}
\end{equation}

A candidate is eligible as chosen if $N_+(v_i)\geq2$ and $N_-(v_i)=0$, and as rejected if $N_-(v_i)\geq2$ and $N_+(v_i)=0$:

\begin{equation}
\begin{aligned}
 \mathcal C_c
 &=\left\{v_i\in\mathcal V_c:
 N_+(v_i)\geq2,\,
 N_-(v_i)=0\right\},\\
 \mathcal R_c
 &=\left\{v_i\in\mathcal V_c:
 N_-(v_i)\geq2,\,
 N_+(v_i)=0\right\}.
\end{aligned}
\label{eq:consensus_filter}
\end{equation}

We form a preference pair only if the group contains exactly one eligible chosen video $v^+$ and one eligible rejected video $v^-$, with $v^+\neq v^-$. All other groups are discarded:

\begin{equation}
\begin{aligned}
 \mathcal D_{\mathrm{pref}}
 =\Big\{&(c,v^+,v^-)\;\Big|\;
 \mathcal C_c=\{v^+\},\\
 &\mathcal R_c=\{v^-\},\;
 v^+\neq v^-\Big\}.
\end{aligned}
\label{eq:preference_dataset}
\end{equation}

\subsection{Training Pipeline}

\subsubsection{Supervised Fine-tuning}

Given a training video $v$ and condition $c$, let $z=\mathcal E(v)$ denote its latent representation. We sample Gaussian noise $\epsilon\sim\mathcal N(0,\mathbf I)$ and a flow-matching time $\tau\in[0,1]$ to construct the interpolated latent and its target velocity:

\begin{equation}
 z_\tau=(1-\tau)z+\tau\epsilon,
 \qquad \mathbf u=\epsilon-z.
 \label{eq:sft_flow_path}
\end{equation}

The velocity field predictor $\mathbf u_\theta$ is trained to estimate $\mathbf u$ from the noisy latent and conditioning information. We define the sample-wise prediction error as:

\begin{equation}
 e_\theta(z,c,\tau,\epsilon)=
 \operatorname{MSE}\big(
 \mathbf u_\theta(z_\tau,\tau,c),\mathbf u
 \big),
 \label{eq:velocity_error}
\end{equation}

The supervised fine-tuning objective is:

\begin{equation}
 \mathcal L_{\mathrm{SFT}}=
 \mathbb E_{(z,c)\sim\mathcal D_{\mathrm{SFT}},\,\tau,\epsilon}
 \big[w(\tau)e_\theta(z,c,\tau,\epsilon)\big],
 \label{eq:sft_loss}
\end{equation}

where $w(\tau)$ denotes the time-dependent training weight. The initial image remains clean and is excluded from the velocity prediction loss.

\subsubsection{Direct Preference Optimization}

We initialize the trainable policy and a frozen reference model from the same SFT checkpoint.\cite{rafailov2023direct,wallace2024diffusion} The preference dataset $\mathcal D_{\mathrm{pref}}$ consists of triples $(c,v^+,v^-)$. We encode the preferred and dispreferred videos into latents $z^+=\mathcal E(v^+)$ and $z^-=\mathcal E(v^-)$.

For each pair, we sample a shared flow-matching time $\tau$ and Gaussian noise $\epsilon$. The frozen reference model uses the same velocity prediction error as the trainable policy:

\begin{equation}
 e_{\mathrm{ref}}(z,c,\tau,\epsilon)=
 \operatorname{MSE}\big(
 \mathbf u_{\mathrm{ref}}(z_\tau,\tau,c),\mathbf u
 \big),
 \label{eq:reference_velocity_error}
\end{equation}

where $\mathbf u_{\mathrm{ref}}$ denotes the velocity prediction of the frozen reference model, and $z_\tau$ and $\mathbf u$ are constructed according to Eq.~\ref{eq:sft_flow_path}.

We define the relative error differences for the preferred and dispreferred samples as:

\begin{equation}
 \Delta_\theta^\pm=
 e_\theta(z^\pm,c,\tau,\epsilon)-
 e_{\mathrm{ref}}(z^\pm,c,\tau,\epsilon),
 \label{eq:dpo_error_difference}
\end{equation}

We optimize the velocity-error preference objective:

\begin{equation}
 \mathcal L_{\mathrm{DPO}}
 =-\mathbb E_{\mathcal D_{\mathrm{pref}},\,\tau,\epsilon}
 \left[
 \log\sigma\left(
 -\beta(\Delta_\theta^+-\Delta_\theta^-)
 \right)
 \right],
 \label{eq:dpo_loss}
\end{equation}

where $\sigma$ is the sigmoid function and $\beta>0$ controls the preference strength. This objective encourages a larger reduction in velocity prediction error for preferred videos than for dispreferred videos, relative to the reference model.

To perform parameter-efficient post-training, we restrict updates to low-rank\cite{hu2022lora} adapters. For each adapted weight matrix, we use:

\begin{equation}
 W=W_{\mathrm{SFT}}+\frac{\alpha}{r_{\mathrm L}}BA,
 \label{eq:lora_update}
\end{equation}

where $A\in\mathbb R^{r_{\mathrm L}\times d_{\mathrm{in}}}$, $B\in\mathbb R^{d_{\mathrm{out}}\times r_{\mathrm L}}$, $r_{\mathrm L}\ll\min(d_{\mathrm{in}},d_{\mathrm{out}})$, and $\alpha$ denotes the adapter scaling factor. Only $A$ and $B$ are optimized; the SFT backbone weights and reference model remain frozen.

\begin{figure*}[t]
    \centering
    \includegraphics[width=\textwidth]{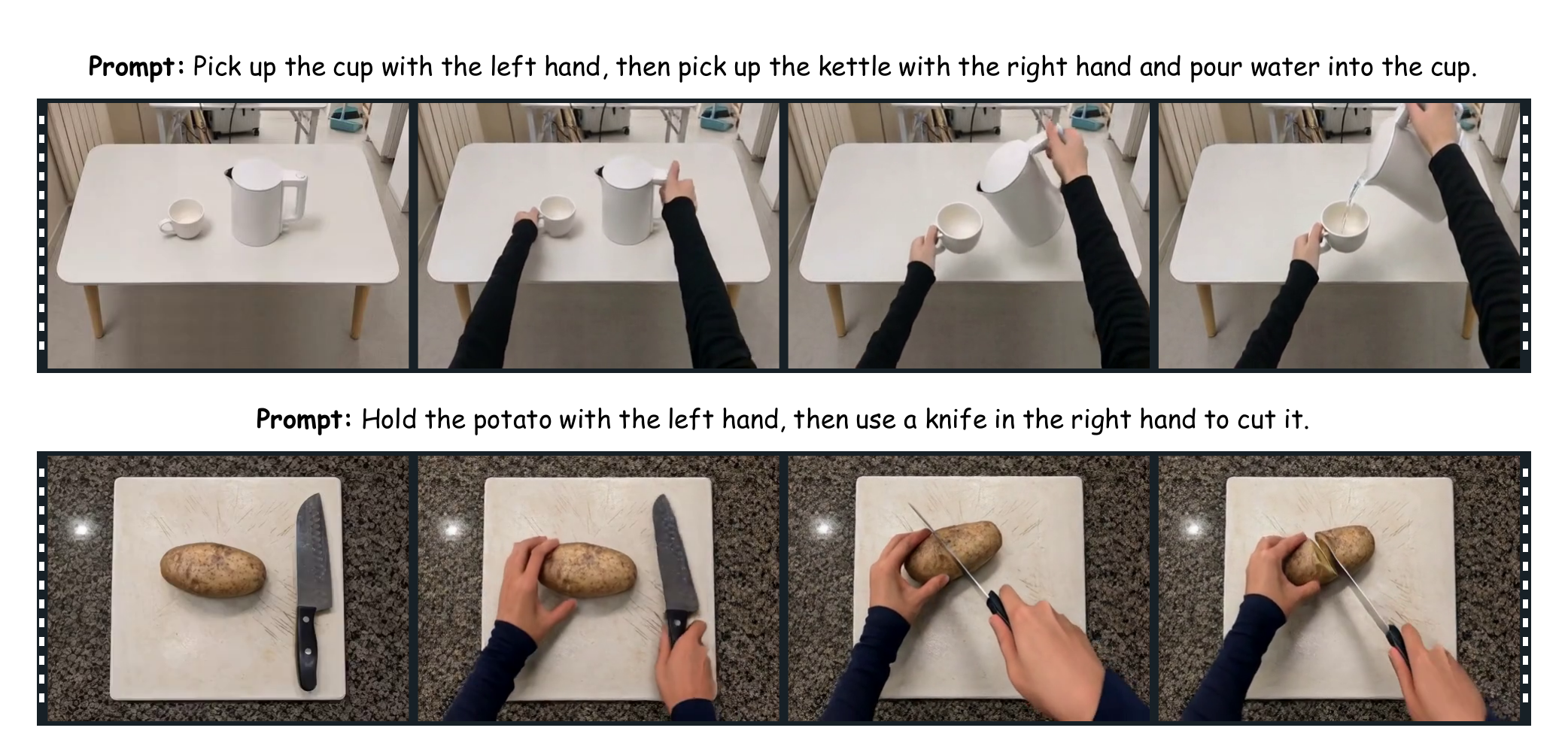}
    \caption{\textbf{In-the-Wild Demonstrations.}
    Qualitative results of our method in diverse real-world scenarios.
    Our method generates visually realistic HOI videos with coherent hand-object motions, plausible interactions, and consistent spatiotemporal dynamics.}
    \label{fig:inthewild}
\end{figure*}
\begin{table*}[t]
    \centering
    \caption{Quantitative comparison on TACO dataset.}
    \label{result_taco}
    \setlength{\tabcolsep}{3pt}
    \renewcommand{\arraystretch}{1.15}
    {%
    \begin{tabular}{lccccccccc}
        \toprule
        Method
        & FVD $\downarrow$
        & SSIM $\uparrow$
        & PSNR $\uparrow$
        & LPIPS $\downarrow$
        & MPJPE $\downarrow$
        & Contact $\uparrow$
        & Coupling $\uparrow$
        & CLIP $\uparrow$
        & R@3 $\uparrow$ \\
        \midrule

        Wan2.2-SFT\cite{wan2025wan}
        & 201.0452 & 0.5180 & 14.5190 & 0.4928 & 39.0253
        & 0.5938 & 0.5806 & 0.2947 & 0.2227 \\

        MIMIC\cite{chen2026mimic}
        & 198.4997 & 0.5195 & 14.5401 & \underline{0.4794} & 38.8650
        & 0.6708 & 0.6138 & 0.2926 & 0.2472 \\

        HVG-3D\cite{chen2026hvg}
        & 193.8961 & 0.5219 & 14.5562 & 0.4822 & 38.2587
        & 0.6803 & \underline{0.6253} & 0.2932 & 0.2406 \\

        HarmoHOI\cite{dang2026harmohoi}
        & 192.9050 & 0.5211 & 14.5558 & 0.4801 & \underline{38.0115}
        & 0.6563 & 0.6241 & 0.2940 & \underline{0.2511} \\

        VideoGPA\cite{du2026videogpa}
        & 191.4962 & \underline{0.5256} & 14.5693 & 0.4845 & 38.1402
        & 0.6639 & 0.5998 & \underline{0.2958} & 0.2464 \\

        PhyGDPO\cite{cai2026phygdpo}
        & \underline{189.8840} & 0.5230 & \underline{14.5845} & 0.4821 & 38.6576
        & \underline{0.7043} & 0.5944 & 0.2932 & 0.2370 \\

        \midrule
        \textbf{MIRA}
        & \textbf{177.4528}
        & \textbf{0.5307}
        & \textbf{14.7618}
        & \textbf{0.4766}
        & \textbf{37.2087}
        & \textbf{0.7396}
        & \textbf{0.6543}
        & \textbf{0.2964}
        & \textbf{0.2559} \\

        \bottomrule
    \end{tabular}%
    }
\end{table*}

\begin{table*}[t]
    \centering
    \caption{Quantitative comparison on ARCTIC dataset.}
    \label{result_arctic}
    \setlength{\tabcolsep}{3pt}
    \renewcommand{\arraystretch}{1.15}
    {%
    \begin{tabular}{lccccccccc}
        \toprule
        Method
        & FVD $\downarrow$
        & SSIM $\uparrow$
        & PSNR $\uparrow$
        & LPIPS $\downarrow$
        & MPJPE $\downarrow$
        & Contact $\uparrow$
        & Coupling $\uparrow$
        & CLIP $\uparrow$
        & R@3 $\uparrow$ \\
        \midrule

        Wan2.2-SFT\cite{wan2025wan}
        & 408.2352 & 0.5078 & 11.6062 & 0.5741
        & 42.9465 & 0.3794 & 0.8645 & 0.2605 & 0.1166 \\

        MIMIC\cite{chen2026mimic}
        & 394.4748 & 0.5196 & 11.7014 & 0.5686
        & 42.6769 & 0.3845 & 0.8732 & 0.2624 & 0.1173 \\

        HVG-3D\cite{chen2026hvg}
        & 396.6385 & 0.5204 & 11.6895 & 0.5598
        & 42.5836 & 0.3796 & 0.8767 & 0.2711 & 0.1183 \\

        HarmoHOI\cite{dang2026harmohoi}
        & \underline{391.0543} & 0.5213 & 11.7875 & 0.5593
        & 42.6158 & 0.3990 & 0.8759 & 0.2707 & 0.1211 \\

        VideoGPA\cite{du2026videogpa}
        & 397.7049 & \underline{0.5379} & 11.7572
        & \underline{0.5501} & \underline{41.3439}
        & 0.4048 & 0.8765 & 0.2702 & \underline{0.1225} \\

        PhyGDPO\cite{cai2026phygdpo}
        & 395.2380 & 0.5296 & \underline{11.8047}
        & 0.5524 & 42.2657 & \underline{0.4065}
        & \underline{0.8810} & \underline{0.2717} & 0.1121 \\

        \midrule
        \textbf{MIRA}
        & \textbf{373.5968}
        & \textbf{0.5394}
        & \textbf{11.9516}
        & \textbf{0.5490}
        & \textbf{41.1613}
        & \textbf{0.4197}
        & \textbf{0.8854}
        & \textbf{0.2782}
        & \textbf{0.1327} \\

        \bottomrule
    \end{tabular}%
    }
\end{table*}

\begin{table*}[t]
    \centering
    \caption{Ablation Study on TACO dataset.}
    \label{abl_taco}
    \setlength{\tabcolsep}{3pt}
    \renewcommand{\arraystretch}{1.15}
    {%
    \begin{tabular}{lccccccccc}
        \toprule
        Method
        & FVD $\downarrow$
        & SSIM $\uparrow$
        & PSNR $\uparrow$
        & LPIPS $\downarrow$
        & MPJPE $\downarrow$
        & Contact $\uparrow$
        & Coupling $\uparrow$
        & CLIP $\uparrow$
        & R@3 $\uparrow$ \\
        \midrule

        w/o 2D RM
        & 184.3976 & 0.5247 & 14.6982 & 0.4785 & 37.4591
        & 0.7188 & 0.6157 & 0.2959 & 0.2454 \\

        w/o 3D RM
        & 182.9543 & 0.5294 & 14.6035 & 0.4815 & 37.8192
        & 0.7194 & 0.6115 & 0.2956 & 0.2278 \\

        w/o 4D RM
        & 181.8702 & 0.5198 & 14.6158 & 0.4837 & 38.0641
        & 0.7106 & 0.6095 & 0.2952 & 0.2396 \\

        w/o RPA
        & 202.9836 & 0.5145 & 14.2407 & 0.4902 & 37.6425
        & 0.7294 & 0.6144 & 0.2955 & 0.2464 \\

        w/o Data Filter
        & 187.2376 & 0.5286 & 14.5835 & 0.4833 & 37.9058
        & 0.7306 & 0.6292 & 0.2960 & 0.2417 \\

        \midrule
        \textbf{MIRA}
        & \textbf{177.4528}
        & \textbf{0.5307}
        & \textbf{14.7618}
        & \textbf{0.4766}
        & \textbf{37.2087}
        & \textbf{0.7396}
        & \textbf{0.6543}
        & \textbf{0.2964}
        & \textbf{0.2559} \\

        \bottomrule
    \end{tabular}%
    }
\end{table*}

\section{Experiments}

We systematically evaluate MIRA on the TACO and ARCTIC datasets to assess its video generation quality and hand-object interaction accuracy. Through quantitative comparisons with representative methods and ablation studies, we investigate the overall performance of our framework and the effectiveness of its key components.

\subsection{Dataset}

\paragraph{TACO Dataset.}
TACO~\cite{taco} is a large-scale bimanual hand-object interaction dataset containing approximately 2,500 sequences of daily tool use and object manipulation, covering diverse actions and object categories. It provides synchronized egocentric and allocentric multi-view videos, along with 3D hand poses and object geometry annotations.

\paragraph{ARCTIC Dataset.}
ARCTIC~\cite{fan2023arctic} focuses on bimanual manipulation and articulated object interactions, containing approximately 2.1 million multi-view RGB frames from 10 subjects interacting with 11 articulated objects. It provides egocentric and allocentric views, along with 3D hand poses, object geometry, and dynamic contact annotations, capturing diverse bimanual manipulations and object state changes.

\subsection{Implementation Details}

For a fair comparison, all methods are trained with a total budget of 480k optimization steps. For RL-based methods, we first perform 384k steps of supervised fine-tuning (SFT) using DiffSynth-Studio to adapt the model to HOI video generation, followed by 96k steps of RL post-training from a shared SFT checkpoint. Other methods follow their respective training strategies under the same total training budget. All videos are standardized to 121 frames at 24 FPS with a resolution of 480 $\times$ 352. Identical sampling steps and guidance scales are used during inference. We allocate 90\% of the data for training and reserve the remaining 10\% for testing. All experiments were conducted using 16 NVIDIA A100 GPUs.

\subsection{Evaluation Metrics}

Following prior work~\cite{manivideo,jiang2023motiongpt,zhang2026photohoi}, we adopt nine metrics, including four general video quality metrics and five HOI-specific metrics.

\paragraph{General Metrics.}
We employ Fréchet Video Distance (FVD) to measure the distributional discrepancy between generated and real videos, Learned Perceptual Image Patch Similarity (LPIPS) to evaluate perceptual differences, Peak Signal-to-Noise Ratio (PSNR) to measure pixel-level reconstruction quality, and Structural Similarity Index Measure (SSIM) to assess image structural similarity.

\paragraph{HOI-specific Metrics.}
We employ Mean Per Joint Position Error (MPJPE) to measure 3D hand joint position errors, Hand-Object Contact Accuracy (Contact) to assess contact accuracy in hand-object interactions, Hand-Object Interaction Motion Coupling (Coupling) to measure motion consistency in hand-object interactions. For Text-to-HOI Alignment, we use CLIP cosine similarity (CLIP) to measure text-to-video semantic similarity and Recall@3 (R@3) to evaluate video-to-text retrieval performance.

\subsection{Main Results}

We conduct quantitative evaluations of MIRA on the TACO and ARCTIC datasets and compare it with several representative methods. Experimental results demonstrate that MIRA achieves state-of-the-art performance across all nine evaluation metrics on both datasets.Detailed results are presented in Tables~\ref{result_taco} and~\ref{result_arctic}.

\subsection{Visualization Results}

\paragraph{Comparison with the Base Model.}
Figure~\ref{fig:teaser} compares videos generated by MIRA and the Wan2.2 base model from the same initial frame. These results illustrate the benefit of unified reward alignment in reducing geometric distortions during dynamic interactions.

\paragraph{In-the-Wild Results.}
Figure~\ref{fig:inthewild} presents robust results in real-world scenarios, where MIRA maintains visual quality, coherent hand-object motion, and consistent object structure, suggesting that the benefits of our post-training extend beyond the benchmark settings.

\subsection{Analysis of Reward Consistency}
We analyze reward consistency on TACO using candidate groups of three generated videos. For the 2D-3D, 2D-4D, and 3D-4D reward pairs, the mean Spearman rank correlations are 0.2864, 0.2530, and 0.2769, respectively, while the preference agreement rates are 62.75\%, 61.31\%, and 62.50\%. These results indicate relatively weak correlations but complementary preference signals across reward dimensions. Our consensus-based filtering retains 71.09\% of the original data for RL post-training, supporting reliable multi-dimensional preference construction.

\subsection{Ablation Study}

We conduct ablation studies on the TACO dataset by individually removing the 2D, 3D, and 4D reward models, rank-based preference assignment (RPA), and data filtering module. As shown in Table~\ref{abl_taco}, removing any component leads to varying degrees of performance degradation across the evaluation metrics, demonstrating the effectiveness of each component.

\paragraph{2D Reward Model (w/o 2D RM).}

The 2D reward evaluates generated videos in terms of visual quality and semantic consistency, focusing on image clarity, fine-grained details, visual realism, and alignment with text descriptions. Removing this reward degrades video generation quality and perceptual fidelity, demonstrating the importance of visual and semantic supervision.

\paragraph{3D Reward Model (w/o 3D RM).}

The 3D reward evaluates cross-view geometric consistency through shared-scene reconstruction and reprojection across sampled views. Removing this reward degrades video generation quality and spatial consistency, highlighting the importance of cross-view geometric supervision for maintaining consistent scene structures.

\paragraph{4D Reward Model (w/o 4D RM).}

The 4D reward primarily evaluates HOI motion consistency through bimanual hand-state trajectories, while also considering dynamic scene-structure consistency. Removing this reward degrades multiple evaluation metrics, with the most pronounced deterioration observed in hand pose accuracy.

\paragraph{Rank-based Preference Assignment (w/o RPA).}

Rank-based preference assignment converts reward evaluations from different dimensions into within-group rankings, reducing the influence of numerical scale differences on preference construction. Removing this strategy leads to noticeable degradation across all nine evaluation metrics, demonstrating its effectiveness in integrating heterogeneous reward signals and providing more reliable preference supervision for RL post-training.

\paragraph{Data Filtering (w/o Data Filter).}

The consensus-based filtering strategy aggregates preference judgments across different reward dimensions to exclude candidate groups with conflicting evaluations or unreliable preferences, thereby constructing high-confidence training data. Removing this strategy degrades generation performance, demonstrating that consensus-based filtering reduces the impact of unreliable preference data on RL post-training.
\section{Conclusion}

In this paper, we present \textbf{MIRA}, an RL post-training framework for improving hand-object interaction video generation. We introduce a unified reward modeling framework that integrates 2D, 3D, and 4D rewards to evaluate visual quality and semantic alignment, 3D geometric and cross-view consistency, and hand-motion and dynamic scene-structure consistency. Furthermore, we develop a rank-based preference assignment strategy and a consensus-based data filtering strategy to construct high-confidence preference pairs for model optimization. Extensive experiments demonstrate that MIRA achieves state-of-the-art performance across all evaluated metrics on multiple datasets, generating videos with improved visual quality, spatial consistency, and physical plausibility. The video generation models enhanced by MIRA could provide high-quality interaction videos for AR/VR, robotic manipulation, and world action models (WAMs), facilitating more realistic virtual interactions and embodied intelligence applications. Future work will explore long-horizon interaction generation and streaming video generation to support more continuous and complex dynamic interaction scenarios.

\clearpage

{
    \small
    \bibliographystyle{ieeenat_fullname}
    \bibliography{main}
}

\end{document}